\documentclass{article}
\usepackage{spconf,amsmath,graphicx}
\usepackage{algorithm,multirow}
\usepackage[noend]{algpseudocode}

\title{Transferable Positive/Negative speech emotion recognition via class-wise adversarial domain adaptation}
\name{Hao Zhou, Ke Chen}
\address{School of Computer Science, The University of Manchester, Manchester, M13 9PL, U.K.}
\begin{document}
%
\maketitle
\begin{abstract}
Speech emotion recognition plays an important role in building more intelligent and human-like agents. Due to the difficulty of collecting speech emotional data, an increasingly popular solution is leveraging a related and rich source corpus to help address the target corpus. However, domain shift between the corpora poses a serious challenge, making domain shift adaptation difficult to function even on the recognition of positive/negative emotions. In this work, we propose class-wise adversarial domain adaptation to address this challenge by reducing the shift for all classes between different corpora. Experiments on the well-known corpora EMODB and Aibo demonstrate that our method is effective even when only a very limited number of target labeled examples are provided.
\end{abstract}
\section{Introduction}
\label{sec:intro}

Speech emotion recognition \cite{el2011survey} has attracted growing interest over the past two decades. It can be applied to many areas. For example, the recognition of positive/negative emotions can help improve the call center service and psychological disease diagnosis. However, it is highly difficult to collect a large volume of speech emotional data in a completely natural environment. Although considerable efforts have been made to build high-quality databases (corpora) of emotional speeches, data scarcity remains a bottleneck. 

An increasingly popular solution to data scarcity is domain shift adaptation or transfer learning \cite{pan2010survey, hassan2013acoustic}. It features leveraging a different but related, and information-rich source domain to improve the performance on the target domain, which we are really interested in but suffers from a lack of information. In the context of speech emotion recognition, a corpus collected in a specific way may be viewed as a domain. There are usually two cases of domain shift adaptation depending on whether label information in the target domain is available: unsupervised domain adaptation (UDA) and supervised domain adaptation (SDA). In either case, the key is eliminating domain shift, i.e. the difference of distributions, between the source and target domains. In speech emotion recognition, domain shift causes the well-known cross-corpora problem that the performance of a recognition system built on one corpus can be degraded significantly when being tested on a different corpus. Consequently, even recognition of positive/negative emotions can be challenging in the cross-corpora setting or in the scenario of domain shift adaptation. Moreover, the problem is even exacerbated when only very limited training data are available in the target domain. In this situation, it even poses a new challenge that the limited training data are insufficient to build up an effective speech emotion recognition system. Hence, the data in different domains/corpora have to be utilized. 

So far there are several approaches \cite{hassan2013acoustic,deng2014autoencoder,deng2014introducing}
developed to address the domain shift issue, but most of them are applied to UDA that requires a great amount of unlabeled target-domain data be accessible during the training stage. This requirement, however, is often difficult to meet considering the difficulty of collecting speech emotional data. On the other hand, SDA can be applied as long as there are a small number of target-domain labeled data available, even if those training data are insufficient to build up an effective recognizer. For instance, a recent method named few-shot adversarial domain adaptation (FADA) \cite{motiian2017few} emerges as a promising SDA approach. The FADA works by aligning distributions across domains via adversarial learning in a discriminative manner. This strategy, however, may encounter a difficulty caused by very high intra-class variability in each corpus considering that the source and target corpora can differ with respect to the speech recording environment, the eliciting way of emotions (natural or acted), the speakers (subjects), the communication language, the pre-defined emotional states, and the emotion annotation scheme, etc. Hence, how to deal with very high intra-class variability in domain shift adaptation becomes an obstacle in transferable speech emotion recognition.

In this paper, we propose the class-wise adversarial domain adaptation (CADA) method to address the intra-class variability issue in domain shift adaptation towards improving the performance on the target domain for the positive/negative emotion recognition. Given a related source domain for which sufficient labeled data are provided, CADA employs adversarial learning in a generative manner to align each class pair between the source and target domains and hence is more effective for knowledge transfer in domain shift adaptation. The experiments on the well-known corpora EMODB and Aibo demonstrate that CADA is effective even when only a very limited number of target labeled examples are provided.

\section{Related work}
\label{sec:related}

Some pioneer works have systematically evaluated cross-corpora speech emotion recognition with a number of high-quality databases \cite{schuller2010cross,eyben2010cross}. These works, unfortunately, do not involve any adaptation techniques for reducing the domain shift. \cite{hassan2013acoustic} first treats the mismatch in emotional data as covariate shift and proposes compensating for that shift by classical importance-weighting at the instance level. At the feature level, some autoencoder-based transfer learning methods \cite{deng2014autoencoder,deng2014introducing} have developed to seek a shared feature representation so that the knowledge can be transferred between the domains. All of these methods, however, are usually applied to unsupervised rather than supervised domain adaptation, which demands a lot of data in target domain that may not be easy to collect in reality.

Regarding supervised domain adaptation in speech emotion recognition, some works have verified that a few labeled data from the target domain can be hugely helpful, but the adaptation is achieved by simple fine-tuning \cite{abdelwahab2015supervised}. A sophisticated technique is by \cite{deng2013sparse}. But still based on autoencoder, this technique needs a relatively large number of target examples for adaptation. Recently adversarial learning \cite{ganin2016domain, tzeng2017adversarial} gains a great popularity on domain shift adaptation. A significant trait about adversarial learning is that instead of directly measuring the similarity of different domains, it introduces a domain discriminator that distinguishes the source from the target domain, and a feature representation is then learned to be domain invariant by fooling the domain discriminator. As the state-of-the-art supervised domain adaptation approach, FADA \cite{motiian2017few} conducts adversarial learning in a discriminative way on generated pairs which mix multi-class training examples by considering the combination of classes in different domains. Although FADA yields good performance in different applications, it does not work well in speech emotion recognition as the method cannot deal with the high intra-class variability effectively in domain shift adaptation. In order to overcome this weakness, our CADA decomposes the domain shift problems on the basis of each common class in the source and target domains, leading to more effective adversarial learning. In addition, by contrast to FADA which employs typical Siamese networks feeding on paired examples, CADA can be implemented easily with a slightly modified multilayer perceptron (MLP).

\section{Our CADA approach}
\label{sec:cada}

Formally, given the source domain $D_s$ and target domain $D_t$ (we use $s$ and $t$ to refer to the source and target domain, respectively), where the source domain follows the distribution $P(X^s, Y^s)$ and the target domain $P(X^t, Y^t)$ (in our context, $X$ denotes the input speech and $Y$ the emotional class), the goal of domain shift adaptation is to learn a classification function $f$ that minimizes the misclassification error $L_y(f(X^t), Y^t)$ by using all the data available in two domains. Under the setting of supervised domain adaptation, $D_s=\{(x^s_i,y^s_i)\}^N_{i=1}$ and $D_t=\{(x^t_i,y^t_i)\}^M_{i=1}$ ($M<<N$). 

A typical domain shift adaptation method usually works under the assumption $P(Y^s|X^s) = P(Y^t|X^t)$. By learning a feature space $\phi$ such that $P(\phi(X^s))=P(\phi(X^t))$, it ideally leads to $P(Y^s|\phi(X^s))=P(Y^t|\phi(X^t))$, which means the classifier can be shared by both domains. However, in speech emotion recognition, the underlying assumption that $P(Y^s|X^s) = P(Y^t|X^t)$ may be less solid because of the high intra-class variability between the source and target domains, and this further affects the learning process for the desired feature space. 

To tackle this weakness, class-wise adversarial domain adaptation (CADA) is aimed at seeking a feature space $\phi$ for $P(\phi(X^s)|y_i)=P(\phi(X^t)|y_i)$, $y_i\in Y$ instead of $P(\phi(X^s))=P(\phi(X^t))$. With the assumption that $P(Y^s)=P(Y^t)$, by Bayesian theory, we wish to have 
\begin{equation}
\begin{split}
    P(y_i|\phi(X^s)) & =\frac{P(\phi(X^s)|y_i)P(y_i)}{\Sigma_i P(\phi(X^s)|y_i)P(y_i)}\\& \approx \frac{P(\phi(X^t)|y_i)P(y_i)}{\Sigma_i P(\phi(X^t)|y_i)P(y_i)}\\&= P(y_i|\phi(X^t))
\end{split}
\end{equation}
where $\approx$ is carried out for domain shift adaptation. 

To integrate adversarial learning and supervised learning into one process, we introduce a domain-class discriminator which can not only distinguish the classes but also distinguish the domains. To illustrate this idea, we take a binary classification task as an example. The modified discriminator (or classifier) classifies any instance into one of four categories: $d_1$ indicating Class 1 from source domain, $d_2$ Class 2 from source domain, $d_3$ Class 1 from target domain, and $d_4$ Class 2 from target domain. In the testing stage, we perform classification with these 4 categories and treat either the prediction $d_1$ or $d_3$ as Class 1, and either $d_2$ or $d_4$ as Class 2. This categorization scheme can be straightforwardly popularized to the cases involving more classes.

\begin{figure}[h!]
\centering
\includegraphics[scale=0.39]{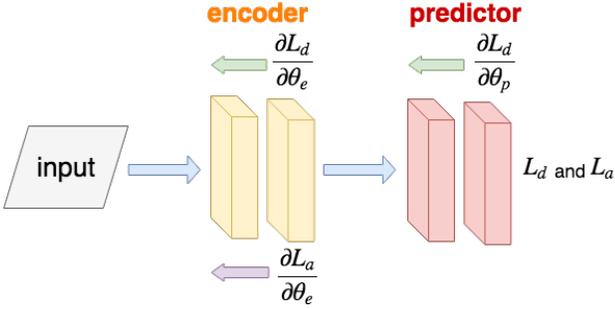}
\caption{The proposed class-wise adversarial learning domain adaptation structure comprises an encoder and a predictor (either may contain more than one hidden layer), parameterized by $\theta_e$ and $\theta_p$ respectively. The training process consists of two stages. In the first stage, both the encoder and predictor are trained based on the loss function $L_d$ defined in Eq.(\ref{cada-loss}). In the next stage, the predictor is fixed and only the encoder is trained based on the loss function $L_a$ defined in Eq.(\ref{lad}). Through the two-stage learning, domain shift is reduced.}
\label{cada}
\end{figure} 

As shown in Fig. \ref{cada}, CADA can be implemented based on an MLP which comprises two components, the feature encoder $G_e$ parameterized by $\theta_e$ and the predictor $G_p$ parameterized by $\theta_p$. Both $\theta_e$ and $\theta_p$ are trained to minimize the typical cross entropy loss function
\begin{equation}
    L_d = -\sum_i^{N+M} d^{x_i}\log G_p(G_e(x_i, \theta _e), \theta _p)
\label{cada-loss}
\end{equation}
where $d^{x_i}$ is the category of $x_i$. Meanwhile, $\theta_e$ is trained to minimize the following loss function
\begin{equation}
\begin{split}
    L_a = & -\{\sum_{x\in X_{d_1}} d_3\log G_p(G_e(x, \theta _e), \theta _p) \\ 
    & + \sum_{x\in X_{d_2}} d_4\log G_p(G_e(x, \theta _e), \theta _p) \\ 
    & + \sum_{x\in X_{d_3}} d_1\log G_p(G_e(x, \theta _e), \theta _p) \\ 
    & + \sum_{x\in X_{d_4}} d_2\log G_p(G_e(x, \theta _e), \theta _p)\}
\end{split}
\label{lad}
\end{equation}
where $X_{d_i}$ denotes all the examples belonging to $d_i$ ($i\in\{1, 2, 3, 4\}$). This loss function is designed to encourage the confusion of the equivalent classes in different domains. Specially, we want the model to believe the examples of certain class in one domain also belong to the equivalent class in the other domain. For instance, the first term on the right side of Eq.(\ref{lad}) suggests that the examples from $d_1$ are also from $d_3$. This principle is applied to all categories we defined. While FADA \cite{motiian2017few} performs adversarial learning on newly-generated data pairs without considering specific class information, minimizing Eq.(\ref{lad}) allows the adversarial learning to operate on each specific common class across the domains, i.e. class-wise adversarial learning. For clarity, the CADA learning process is summarized in Algorithm \ref{al-cada}.

\begin{algorithm}
\caption{CADA learning algorithm}\label{al-cada}
\begin{algorithmic}[1]
\State Initialize $\theta_e$ and $\theta_p$ randomly
\State Re-label training examples in both source and target domains in terms of $d_i$, $i\in\{1,2,3,4\}$
\State \textbf{while} not convergent \textbf{do}
\State \ \ \ \ Update $\theta_e$ and $\theta_p$ by minimizing Eq.(\ref{cada-loss}).
\State \ \ \ \ Update $\theta_e$ by minimizing Eq.(\ref{lad}).
\State \textbf{end while}
\end{algorithmic}
\end{algorithm}

\section{Experiment}
\label{sec:print}

\subsection{Datasets}
We evaluate our proposed CADA with the two well-known speech emotion datasets Aibo \cite{steidl2009automatic} and EMODB \cite{burkhardt2005database} for positive/negative emotion recognition. Despite the same language (German), these two corpora are collected in very different ways. They differ with respect to, at least, the recording environments, speakers (adults or children), the eliciting way of emotions (acted or spontaneous), and the pre-defined emotion class stereotypes. These factors can generate both large domain shift between the two corpora and high intra-class variability for each corpus. Considering the large size of Aibo, we treat the two parts, Aibo-Ohm and Aibo-Mont, separately in our experiments. Details on these three corpora are listed in Table \ref{new-cate}.

\begin{table}[]
\centering
\caption{Datasets}
\label{new-cate}
\resizebox{\columnwidth}{!}{
\begin{tabular}{|l|l|l|l|}
\hline
Corpus   & Type    & Speakers & Class and Size                                                                                                    \\ \hline \hline
EMODB    & acted   & 10 adults       & \begin{tabular}[c]{@{}l@{}}Negative (anger, sadness, etc.): 385\\ Positive (happiness, neutral): 150\end{tabular} \\ \hline
Aibo-Ohm  & spontaneous & 26 children       & \begin{tabular}[c]{@{}l@{}}Negative (angry, touchy, etc.): 3358\\ Positive (joyful, neutral, etc.): 6601\end{tabular}                                    \\ \hline
Aibo-Mont & spontaneous & 25 children       & \begin{tabular}[c]{@{}l@{}}Negative (angry, touchy, etc.): 2465\\ Positive (joyful, neutral, etc.): 5792\end{tabular}                                    \\ \hline
\end{tabular}}
\end{table}

\subsection{Experimental Settings}
To simulate the cross-corpora setting, we use one corpus as the source domain and a different corpus as the target domain. Since both Aibo-Ohm and Aibo-Mont have much more data than EMODB, we consider Aibo-Ohm or Aibo-Mont as the source domain, and EMODB as the target domain, which is consistent with the general experiences to use an information-rich domain as the source. For the transferable positive/negative emotion recognition, the goal is to achieve the best performance on the target domain with few labeled data in the target domain. The performance is measured by unweighted accuracy (UA); i.e. the accuracy per class averaged by the class number.

We further set the following baselines for comparison: 1) \textit{all-source}: we test using the trained source model without any target information (without any adaptation); 2) \textit{all-target}: we use the entire target dataset for traditional supervised learning; 3) \textit{label-target}: we make use of only a few labeled target data (no source domain knowledge used), and this baseline is only introduced when there are over 10 examples per class in the target domain. We use the 5-fold cross validation (under the speaker-dependent condition) where 10\% training examples are preserved for early stopping, and the mean and standard deviation of UA are reported. In addition, we consider two typical adaption methods in our comparison: fine-tuning \cite{abdelwahab2015supervised}, an effective method in speech emotion recognition for transfer learning (with an MLP in our experiment); and FADA \cite{motiian2017few}, a state-of-the-art supervised domain shift adaptation method. For all the adaptation methods, the used target-domain examples are randomly selected from random speakers with the same setting. 20 trials have been conducted for each adaptation method in our experiment and the mean and the standard deviation of UA are reported. 

For preprocessing and feature extraction, we use GeMAPS (62 features) \cite{eyben2016geneva} which has shown a comparable discriminative power as some large standard feature sets but with a much smaller size. We perform normalization by standardizing the feature values to the range $[-1, 1]$. Model selection suggests that an MLP of one hidden layer of 256 ReLu units is a proper base model and our CADA is implemented based on this model as well (c.f. Fig. \ref{cada}). The model is constructed with TensorFlow and trained with the mini-batch of size 64 and the default learning rate via AdamOptimizer.

\subsection{Results Analysis}

Tables \ref{re-oe} and \ref{re-me} report the performance when very few examples (from 2 to 12 with interval of 2) in the target domain are used. Due to the nature of supervised domain adaptation, the effectiveness of adaptation highly depends on the informativeness of the labeled target data which are used in the training process. That explains the relatively large standard deviation in two tables. All the adaptation methods achieve better performance than the baseline \textit{all-source}, suggesting that domain shift adaptation is effective. It is clearly seen from Tables 2 and 3 that our CADA outperforms two state-of-the-art adaptation methods and FADA is inferior to others. 

\begin{figure}[h!]
\centering
\includegraphics[scale=0.45]{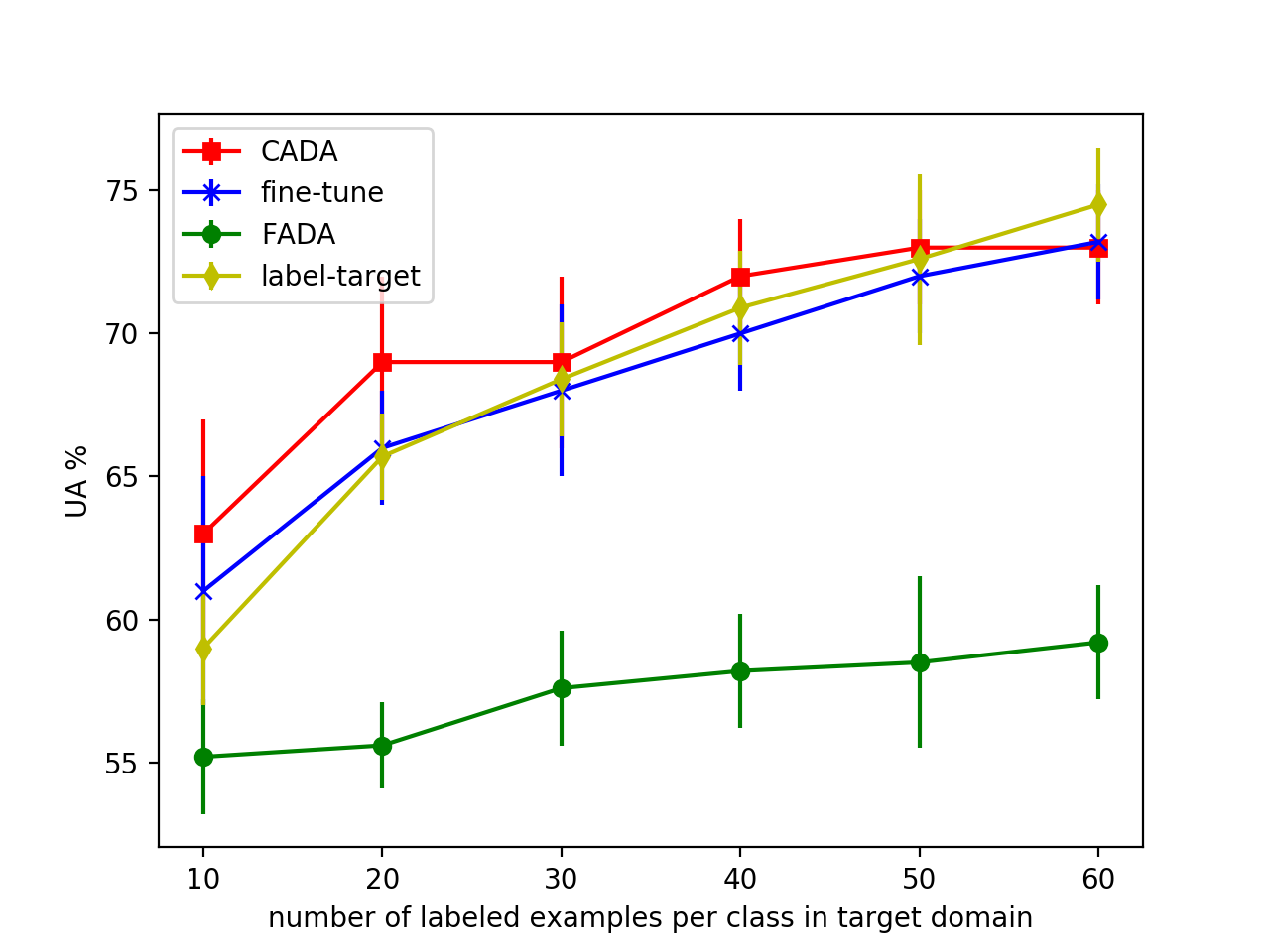}
\caption{Results when using Aibo-Ohm as source domain}
\label{foe}
\end{figure}

\begin{figure}[h!]
\centering
\includegraphics[scale=0.45]{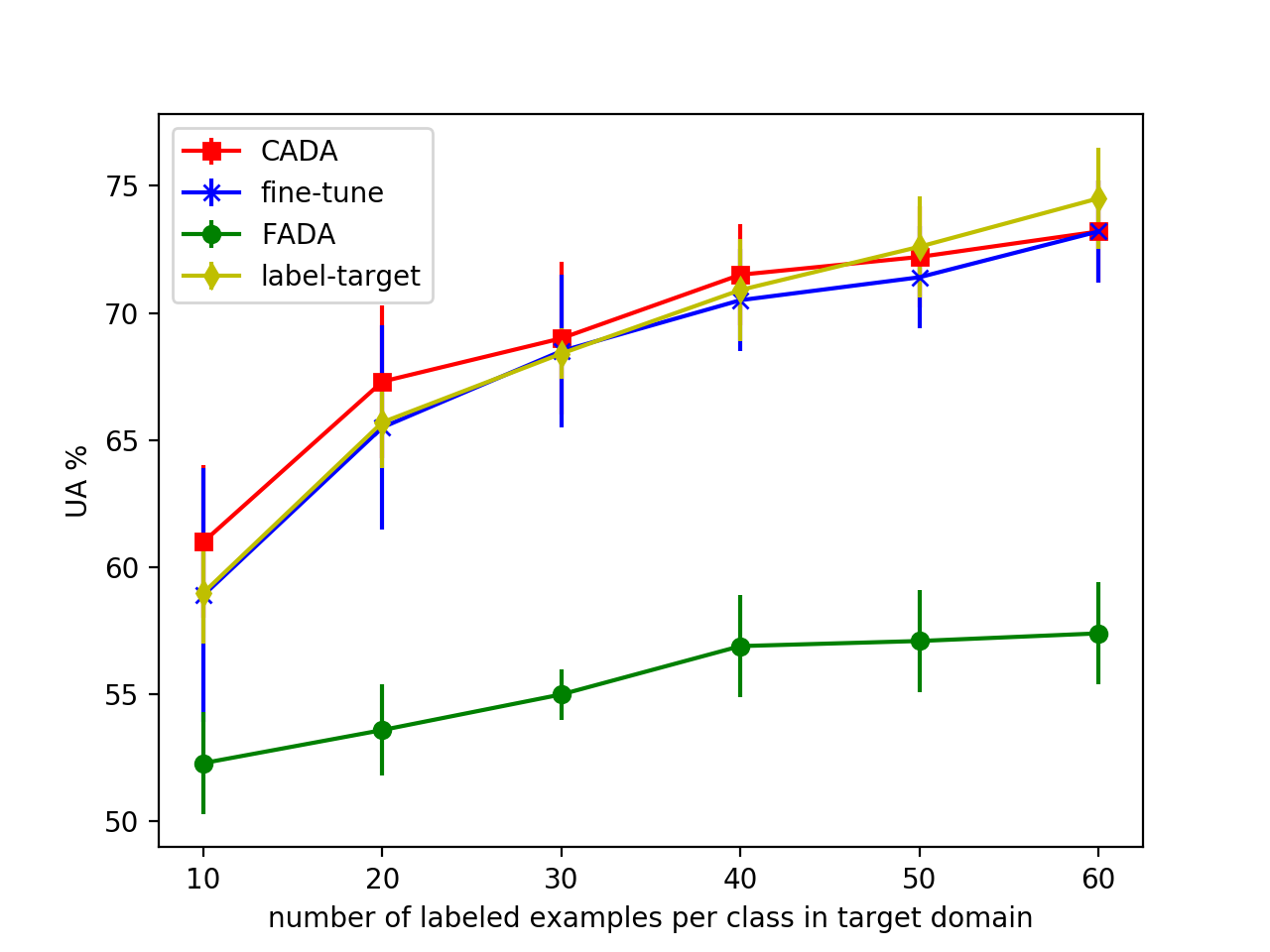}
\caption{Results when using Aibo-Mont as source domain}
\label{fme}
\end{figure}

Figs. \ref{foe} and \ref{fme} show the performance of different adaptation methods and the MLP with the \textit{label-target} setting by using different numbers of target labeled data. It is evident from Figs. \ref{foe} and \ref{fme} that the performance of those adaptation methods is better than that of the \textit{label-target} setting when there are a few labeled data in the target domain. In particular, our CADA always outperforms other adaptation methods and the \textit{label-target} setting up to 50/40 labeled examples in the target domain.

\begin{table}[]
\centering
\caption{Results when using Aibo-Ohm as source domain. Two baselines are \textit{all-source}: 55$\pm$2, and \textit{all-target}: 81$\pm$3.}
\label{re-oe}
\resizebox{\columnwidth}{!}{
\begin{tabular}{|l|l|l|l|l|l|l|}
\hline
Examples  & 2    & 4    & 6    & 8    & 10   & 12   \\ \hline \hline
fine-tune & 55$\pm$4 & 57$\pm$4 & 59$\pm$4 & 60$\pm$5 & 61$\pm$4 & 62$\pm$4 \\ \hline
FADA      & 54$\pm$1 & 54$\pm$2 & 54$\pm$1 & 55$\pm$2 & 55$\pm$1 & 55$\pm$2 \\ \hline
CADA      & \textbf{58$\pm$4} & \textbf{59$\pm$4 }& \textbf{62$\pm$3} & \textbf{63$\pm$4} & \textbf{63$\pm$4} & \textbf{64$\pm$3} \\ \hline
\end{tabular}}
\end{table}

\begin{table}[]
\centering
\caption{Results when using Aibo-Mont as source domain. Two baselines are \textit{all-source}: 51$\pm$1, and \textit{all-target}: 81$\pm$3.}
\label{re-me}
\resizebox{\columnwidth}{!}{
\begin{tabular}{|l|l|l|l|l|l|l|}
\hline
Examples  & 2    & 4    & 6    & 8    & 10   & 12   \\ \hline \hline
fine-tune & 55$\pm$4 & 55$\pm$4 & 57$\pm$5 & 58$\pm$4 & 59$\pm$5 & 61$\pm$4 \\ \hline
FADA      & 50$\pm$1 & 50$\pm$2 & 51$\pm$2 & 52$\pm$2 & 52$\pm$2 & 53$\pm$1 \\ \hline
CADA      & \textbf{57$\pm$3} & \textbf{59$\pm$3} & \textbf{60$\pm$3} & \textbf{60$\pm$4} & \textbf{61$\pm$3} & \textbf{63$\pm$4} \\ \hline
\end{tabular}}
\end{table}

%

\section{Conclusion}
\label{sec:conc}

We have presented a new domain shift adaptation method, CADA, for transferable speech emotion recognition. Our experiments demonstrate that CADA is more effective than the direct use of those training examples in the target domain to build up a system and other adaptation methods when there are few training examples in the target domain. Although CADA was only evaluated on the positive/negative emotion recognition, it is straightforward to be extended to multi-class speech emotion recognition across multiple corpora and other application areas. In our ongoing work, we are going to address the computational issues in multi-class adaptation and extend CADA to unsupervised domain shift adaptation.


\vfill\pagebreak

\bibliographystyle{IEEEbib}
\bibliography{strings,refs}

\end{document}